%% file: main.tex
\documentclass[letterpaper, 10 pt, conference]{ieeeconf}  %

\IEEEoverridecommandlockouts                              %

\usepackage[utf8]{inputenc}
\usepackage[T1]{fontenc}
\usepackage[english]{babel}
\usepackage[colorlinks]{hyperref}
\usepackage{url}
\usepackage{amsfonts}
\usepackage{nicefrac}
\usepackage{microtype}
\usepackage{algorithm}
\usepackage[noend]{algorithmic}
\usepackage{wrapfig}
\usepackage{amssymb}
\usepackage[normalem]{ulem}
\usepackage{mathtools}
\usepackage{subcaption}
\usepackage{ragged2e}
\usepackage{booktabs}
\usepackage{tikz}
\usepackage{etoolbox}
\usepackage{xspace}
\usepackage{dsfont}
\usepackage{xpatch}
\usepackage{enumerate}
\usepackage{xstring}
\usepackage{bbm}
\usepackage{setspace}
\usepackage{tabularx}
\usepackage{graphicx}
\usepackage{xcolor}

\usepackage[useregional=numeric]{datetime2}
\usepackage{cite}

\title{\LARGE \bf
Batch Exploration with Examples for \\ Scalable Robotic Reinforcement Learning
}

\author{
  Annie S. Chen$^*$, HyunJi Nam$^*$, Suraj Nair$^*$, Chelsea Finn\\
  Stanford University\\
   \texttt{\{asc8, hjnam, surajn\}@stanford.edu} \\
  \scriptsize{* equal contribution} 
}

\newcommand{\algoAC}{BEE}
\newcommand{\data}{$\mathcal{D}$}

\newcommand{\reward}{$\mathcal{R}_{exp}$}

\definecolor{mygreen}{rgb}{0.1, 0.6, 0.1}
\newcommand{\rev}{}
\newcommand{\revf}{}

\DeclareMathOperator\supp{supp}

\begin{document}

\maketitle
\thispagestyle{empty}
\pagestyle{empty}

\begin{abstract}
\input{sections/abstract}
\end{abstract}

\input{sections/intro}
\input{sections/related}

\input{sections/prelim}
\input{sections/method}
\input{sections/experiments}

\input{sections/future_work}

\section*{Acknowledgment}

The authors would like to thank Tianhe Yu for help setting up the simulated environments,
as well as Ashwin Balakrishna and members of the IRIS lab for valuable discussions.
This work was supported in part by Schmidt Futures, by ONR grant N00014-20-1-2675, and by an NSF GRFP. Chelsea Finn is a CIFAR Fellow in the Learning in Machines \& Brains program.

\bibliographystyle{IEEEtran}
\bibliography{bee}

\appendices
\input{sections/appendix}

\end{document}

%% file: sections/abstract.tex
Learning from diverse offline datasets is a promising path towards learning general purpose robotic agents.
However, a core challenge in this paradigm lies in collecting large amounts of meaningful data, while not depending on a human in the loop for data collection.
One way to address this challenge is through \emph{task-agnostic exploration}, where an agent attempts to explore without a task-specific reward function, and collect data that can be useful for any subsequent task.
While these approaches have shown some promise in simple domains, they often struggle to explore the relevant regions of the state space in more challenging settings, such as vision-based robotic manipulation. This challenge stems from an objective that encourages exploring \emph{everything} in a potentially vast state space.
To mitigate this challenge, we propose to focus exploration on the important parts of the state space using \emph{weak human supervision}.
Concretely, we propose an exploration technique, Batch Exploration with Examples (\algoAC),
that explores relevant regions of the state-space, guided by a modest number of human-provided images of important states.
These human-provided images only need to be provided once at the beginning of data collection
and can be acquired in a matter of minutes, 
allowing us to scalably collect diverse datasets, which can then be combined with any batch RL algorithm.
We find that \algoAC~is able to tackle challenging vision-based manipulation tasks both in simulation and on a real Franka Emika Panda robot, and observe that compared to task-agnostic and weakly-supervised exploration techniques,
it (1) interacts more than twice as often with relevant objects, and (2) improves subsequent task performance when used in conjunction with offline RL. 

%% file: sections/intro.tex
\section{Introduction}

Learning from large and diverse datasets is a paradigm that has seen remarkable success in several domains, from computer vision \cite{imagenet_cvpr09} to natural language \cite{bert}, with favorable properties like broad generalization. 
Towards extending this success to robotics, we aim to study how we can acquire large amounts of useful robotic interaction data, a problem that remains a significant challenge.
On one hand, having humans explicitly collect meaningful interaction, e.g. through teleoperation \cite{mandlekar2018roboturk}, can be difficult to do at scale. On the other hand, while random exploration techniques can be run at a much larger scale \cite{dasari2019robonet}, they collect lower quality interaction with the environment due to the lack of human supervision.

One general approach to this challenge is to use task-agnostic exploration \cite{bellemare2016unifying,pathakICMl17curiosity, pathak2019self, sekar2020planning}
which leverages some form of intrinsic reward to meaningfully explore in an unsupervised and scalable manner.
While these approaches have been successful in video games and simulated control domains, they can struggle with the requirement of exploring \emph{everything} in real, high-dimensional scenes, such as those in vision-based robotic manipulation problems.
In such settings, there will often be certain regions of the state space that are more important to explore than others. For example, in the case of robotic manipulation, exploring the interactions between the end-effector and objects is likely more important than exploring all possible arm configurations. While one could build heuristics into the algorithm to bias it away from irrelevant exploration (e.g. penalizing videos with only arm motion), such approaches require designing a potentially complex heuristic for each use case. Rather, we aim to provide a general framework for guided exploration, which leverages easy to collect supervision and which can, in principle, be applied in any domain.

\begin{figure}%
    \centering
    \vspace{0.1cm}
    \includegraphics[width=0.99\linewidth]{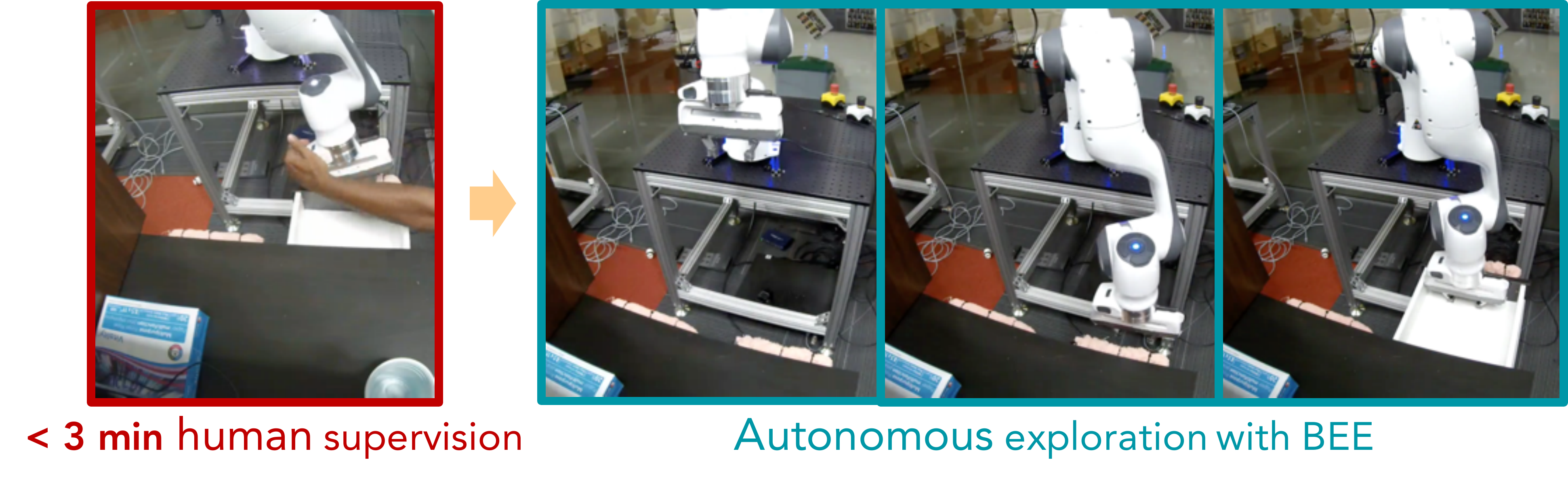}
    \vspace{-0.6cm}
    \caption{\small \textbf{\rev{Batch Exploration with Examples.}} 
    Using just a few minutes of human supervision, our method is able to guide robotic exploration towards more meaningful interactions, such as interacting with a small drawer on a desk. }
    \vspace{-0.4cm}
    \label{teaser}
\end{figure}

Our key insight is that by leveraging some \emph{weak human supervision}, we can allow the agent to focus on semantically relevant parts of the state space, greatly accelerating the collection of useful data. Specifically, a human can communicate a prior over relevant states by providing a handful of examples of ``interesting'' or ``meaningful'' states ahead of time, which a learning-based agent can then use to guide their exploration. Moreover, such human-provided examples can be acquired regardless of domain or use case, enabling much greater flexibility than a domain-specific heuristic.

Concretely, our main contribution is a batch exploration framework, \textbf{B}atch \textbf{E}xploration with \textbf{E}xamples (\algoAC), which leverages weak supervision to efficiently explore and can enable scalable collection of robotic datasets (Figure~\ref{teaser}).
\algoAC~starts with a modest number of examples of relevant states provided by a human, which only takes a few minutes to collect.
\algoAC~then learns to estimate whether a state is relevant or not, and explores around states which it estimates are relevant. It selects exploratory actions via model-based RL, which learns a model of the environment online and plans actions under the model to explore. 
We observe that \algoAC~is able to explore effectively in challenging high-dimensional robotic environments, and unlike standard task agnostic exploration techniques, is able to guide its exploration towards relevant states. Across a range of simulated vision-based manipulation tasks, \algoAC~interacts more than twice as often with relevant objects than prior state-of-the-art unsupervised and weakly supervised exploration methods, and as a result collects higher-quality data, enabling better subsequent, or ``downstream'', task performance. Additionally, by using both weak human supervision and offline model-based RL, \algoAC~can be applied to hard exploration problems from pixels on a \rev{Franka Emika Panda} robot.

%% file: sections/related.tex
\section{Related Work}

Learning from diverse offline datasets has shown promise as a technique for learning robot policies that can generalize to unseen tasks, objects, and domains \cite{finn2017deep, pinto2016supersizing, zeng2018learning,  ebert2018visual, gupta2018robot, kalashnikov2018qtopt, dasari2019robonet, cabi2019scaling, mandlekar2019iris}.
However collecting such large and diverse datasets in robotics remains an open and challenging problem.

A vast number of prior works have collected datasets for robotic learning under a range of problem settings and supervision schemes. One class of approaches uses humans in the loop and collects datasets of task demonstrations via teleoperation \cite{mandlekar2018roboturk, cabi2019scaling} or kinesthetic teaching \cite{sharma2018multiple, xie2019improvisation}. 
While these methods can produce useful data, they are difficult to deploy at scale, across diverse tasks and environments. Alternatively, many other works have explored collecting large robotic datasets without humans in the loop for tasks like object re-positioning \cite{finn2017deep, ebert2018visual, dasari2019robonet}, pushing \cite{yu2016more, agrawal2016learning} and grasping \cite{pinto2016supersizing, Chebotar2016BiGSBG, levine2018learning}.
While these present a scalable approach to data collection, the unsupervised nature of the exploration policy results in only a small portion of the data containing meaningful  interactions. While heuristics and scripted policies like those employed in grasping can enable more meaningful interactions, designing them for a broad range of tasks can require significant engineering effort.

One way to maintain the scalability of random exploration, but acquire more relevant interaction, is to have an agent learn to explore under an intrinsic reward signal, which is task-agnostic but encourages more meaningful interaction. 
These intrinsic rewards come in many forms, including approaches that optimize for visiting novel states \cite{bellemare2016unifying, tang2017exploration, burda2018exploration, ecoffet2019go}, the learning progress of the agent \cite{Oudeyer_LP, oudeyer2018computational}, model uncertainty \cite{schmidhuber_formal, houthooft2016vime, pathak2019self, sekar2020planning}, information gain \cite{houthooft2016vime}, auxiliary tasks \cite{riedmiller2018learning}, generating and reaching goals \cite{pong2019skew, chen2020skewexplore}, and state distribution matching \cite{lee2019efficient}. 
Additionally, a number of these approaches \cite{riedmiller2018learning, pong2019skew, chen2020skewexplore} have been demonstrated on real robotics problems.
\revf{However, one challenge with exploring using only intrinsic rewards is  having to explore \emph{everything} about a vast state space when only some portion of it is relevant.}
We aim to mitigate this challenge by introducing weak supervision into the exploration problem, which we observe empirically yields much more useful exploration than task-agnostic strategies.

A seemingly obvious approach to incorporating supervision into the exploration problem is to include a task-specific extrinsic reward function which is then combined with the exploration objective. In fact, most applications of intrinsic motivation in RL do exactly this, and treat the intrinsic reward as an additional reward bonus. Other works also leverage more complex approaches to combining value functions and exploration \cite{osband2019deep, zhang2020automatic, simmons-edler2020qxplore}. Unlike these works, we aim to not rely on any supervision in the loop of RL, as is needed when providing a reward function online.
Like this work, some prior works have explored how out-of-the-loop weak supervision can be leveraged to acquire better exploratory behavior, ranging from demonstrations \cite{piot2014boosted, hussenot2020show}, binary labels about state factors of variation \cite{lee2020weakly}, and semantic object labels \cite{chaplot2020semantic} to accelerate exploration.
Alternatively, our proposed supervision can be collected in minutes and leads to efficient exploration in real visual scenes of robot manipulation. \revf{While we focus on robot manipulation, recent work \cite{koreitem2020oneshot} also explores using human-provided images to guide marine robot navigation.}

Our method draws inspiration from prior work on reward learning \cite{fu2018variational, singh2019end} and adversarial imitation learning \cite{torabi2018generative}. These approaches aim to tackle the \emph{task-specification problem} by learning a discriminator over human provided goal state images or demonstrations, which is used to acquire a reward function. In contrast, our work focuses on how to incorporate scalable sources of supervision into robotic exploration and data collection. We show that an ensemble of such discriminators can be used to guide exploration, and this data can easily be used with any offline reinforcement learning algorithm. By considering the two-stage batch exploration + batch reinforcement learning approach, our work depends far less on the accuracy of the specific discriminators used during data collection, and can potentially learn multiple downstream tasks from a single dataset.

%% file: sections/prelim.tex
\section{The Batch Exploration + Batch RL Framework}

\begin{figure*}
\centerline{\includegraphics[width=0.9\linewidth]{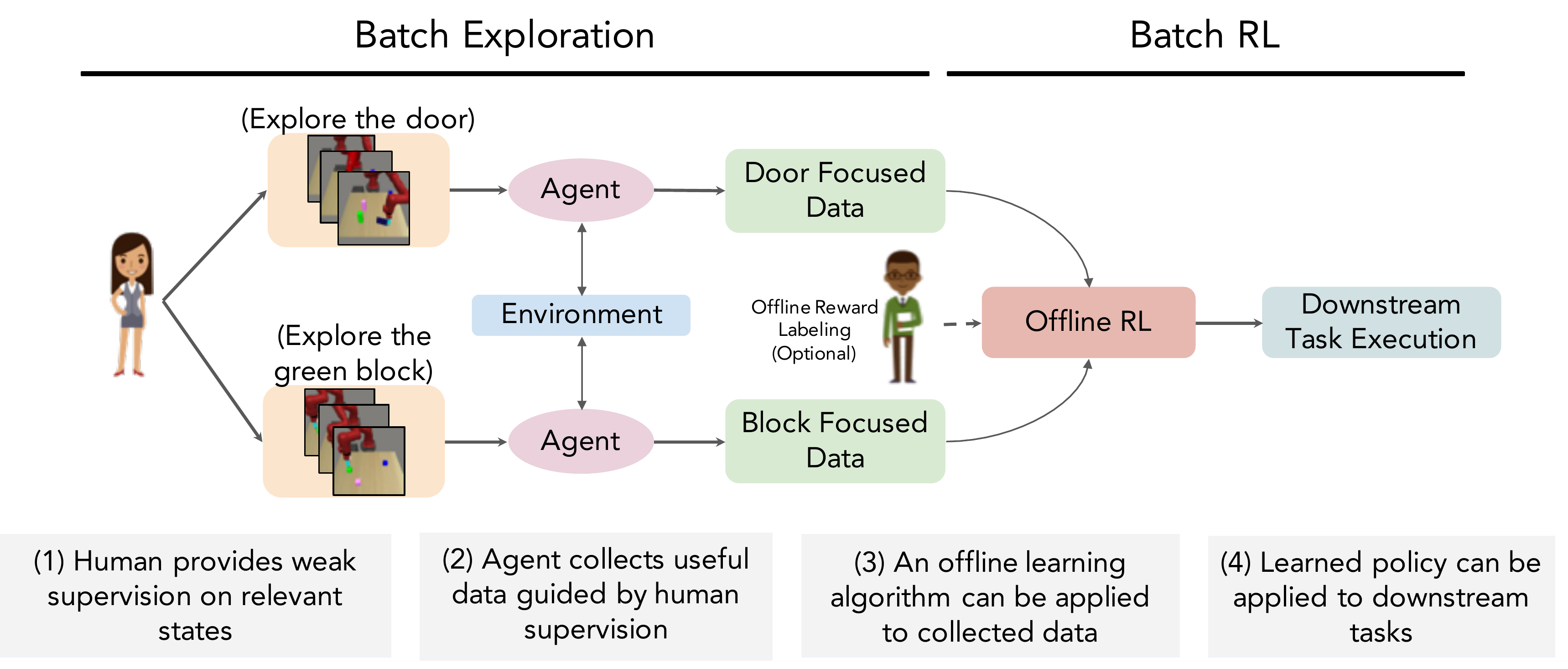}}
\vspace{-0.2cm}
\caption{\small{\textbf{The Batch Exploration + Batch RL Framework}. \rev{Guided by human provided weak supervision, the agent can collect useful data autonomously. Offline RL (either goal-directed or optionally using offline reward labeling) can then be run on this data and the learned policy can be applied to subsequent tasks. Depending on the type of weak supervision provided by the human, the first step can be completed in a matter of minutes, 
and data can be collected and learned from unsupervised and at scale.}
} }

\vspace{-0.3cm}
\label{batch_exp_overview}
\end{figure*}

We begin by describing the batch exploration + batch reinforcement learning framework, with the goal of a \emph{scalable}, data-driven approach to robotic learning,
illustrated in Figure~\ref{batch_exp_overview}. 
\textbf{First}, in the batch exploration phase, a human provides some weak supervision to the agent to indicate what regions of the state space it should explore. \textbf{Second},  also in the batch exploration phase, the agent uses the supervision to guide its exploration to collect and store the relevant data \emph{without needing a human in the loop}. \textbf{Third}, in the batch reinforcement learning phase, the collected datasets can be used with an offline RL algorithm
to learn a policy or model. 
\rev{In principle,} this offline RL phase either can use self-supervised RL techniques (e.g. goal-conditioned model-free RL \cite{Andrychowicz2017HindsightER} or visual foresight \cite{finn2017deep, ebert2018visual}) or can label the offline dataset with rewards and then use standard batch RL algorithms \cite{Levine2020OfflineRL}. \textbf{Lastly}, the policy can then be applied to any  tasks for which the initial guidance was relevant. 
We next describe each of these steps in detail.

\noindent \textbf{Batch Exploration.} During the batch exploration phase, let the agent be exploring in a fixed-horizon controlled Markov process (CMP) $\mathcal{M}$
defined by the tuple $\mathcal{M} = (\mathcal{S}, \mathcal{A}, p, \mu, T)$,
where $\mathcal{S}$ is the state space, $\mathcal{A}$ is the action space, $p(s_{t+1} | s_t, a_t)$ represents the stochastic environment dynamics, $\mu(s_0)$ represents the initial state distribution, and $T$ denotes the episode horizon. Additionally, let $\mathcal{S}^* \subset \mathcal{S}$ represent a subset of the state space which is relevant to explore. Formally, for a set of downstream tasks $\mathcal{T}$, let the relevant states $\mathcal{S}^*$ correspond to the set of all states that have nonzero support under the state marginal distribution $\rho_{\pi_j^*}(s)$ of the optimal policy $\pi^*_j$ for all possible downstream tasks $j \in \mathcal{T}$, that is:
\begin{equation}
\begin{aligned}
\mathcal{S}^* = \bigcup_{j \in \mathcal{T}} \supp \rho_{\pi_j^*}(s) ~~~\text{where}~~~~~~\\
\end{aligned}
\end{equation}
\vspace{-0.5cm}
$$\rho_{\pi} (s) = \mathbb{E}_{s_0 \sim \mu, s_{t+1} \sim p(\cdot \mid s_t, a_t), a_t \sim \pi (\cdot \mid s_t) }\left[\frac{1}{T} \sum_{t=1}^T \mathds{1}\{s_t = s\}\right].$$
    \label{eq1}
In the first stage of batch exploration, a human provides some context information $\mathcal{C}$ to guide the agent towards exploring the relevant states. \rev{Broadly speaking, this context information can take many forms, ranging from a demonstration video to a language description of the relevant states.}  In this work, we consider the case where it is a set of $K$ states
$[\bar{s}_1, ..., \bar{s}_K]$ \rev{that are representative of} $\mathcal{S^*}$. 
In the second stage of batch exploration, the agent aims to learn an exploration policy $a_t \sim \pi_{exp}(\cdot \mid s_t, \mathcal{C})$ conditioned on the human-provided context to maximize an exploratory reward $\mathcal{R}_{exp}(s_t, \mathcal{C})$, which gives high reward for visiting states in $\mathcal{S}^*$, that is find $\pi_{exp}$ which maximizes the expected exploratory reward
\begin{equation}
    \max_{\pi_{exp}} \mathbb{E}_{s_{t+1} \sim p(\cdot \mid s_t, a_t), a_t \sim \pi_{exp}(\cdot \mid s_t, \mathcal{C}), s_0 \sim \mu} [\mathcal{R}_{exp}(s_t, \mathcal{C})].
\end{equation}

The exact reward function implementation will depend on the form of the context $\mathcal{C}$, which we discuss further in Section~\ref{method}.
The policy $\pi_{exp}$ is trained online for $N$ transitions while periodically updating on batches sampled from the memory, 
and after the $N$ transitions all of its collected experience is combined to form the resulting dataset \data~containing tuples $[(s_t, a_t, s_{t+1})_1, ..., (s_t, a_t, s_{t+1})_N]$. This complete dataset can then be used for batch RL.

\noindent \textbf{Batch Reinforcement Learning.} 
Given the dataset \data,
collected by the agent, a number of offline RL approaches can be applied \rev{ranging from goal-conditioned policy learning to labeling some portion of the data with rewards offline and performing standard offline RL.
As described in Section \ref{exps}, we will use a model-based planning approach with goal images or learned reward models. 

%% file: sections/method.tex
\section{Batch Exploration with Examples (\algoAC)}
\label{method}

While a number of works have studied the problem of batch RL \cite{Levine2020OfflineRL},
fewer works have studied the batch exploration phase of our framework, which we focus on here. Our proposed approach, batch exploration with examples (\algoAC) receives weak supervision via examples of relevant states, and subsequently aims to collect data around such relevant states. This latter step is difficult, as it requires the algorithm to determine (a) whether a state (in our case an image) is relevant to explore, and (b) how to select actions to reach these states. 
In this section we describe how the supervision is provided, how \algoAC~determines whether states are relevant, and how we can efficiently learn to reach these relevant states via planning with a learned dynamics model. The complete algorithm is overviewed in Algorithm 1.

\subsection{Acquiring Human Supervision}
The first step of \algoAC~is collecting weak supervision to guide exploration. In this work this supervision comes in the form of a handful of relevant states $[\bar{s}_1, ..., \bar{s}_K]$ \rev{that are representative of} $\mathcal{S^*}$.
\rev{For example, if the relevant region of the state space involves arranging cutlery, these relevant states would consist of images of the robot interacting with forks and knives. Since these images do not need to specify precise states, and are simply meant to guide the robot's exploration, they can be collected in a matter of minutes by manually placing the robot and objects into relevant configurations}\revf{, at least in the case of a manipulator mounted on a tabletop.}
See Figure \ref{fig:disc} for examples states of exploring a desk drawer.

\rev{
How should a human go about providing example states to best guide exploration? 
We find that \algoAC~performs most effectively when the human provides a highly diverse set of example states which fall within $\mathcal{S}^*$. A more diverse set of states allows \algoAC~to generalize more effectively, and better learn the underlying exploration goal. For example, in the example of arranging cutlery, \algoAC~would benefit from human-provided states which show the robot interacting with forks, spoons, and knives in a number of different ways, and from this diversity can learn to explore all cutlery interactions. 
}
\input{sections/algoblock}

\subsection{Exploring with \algoAC}
Online exploration with \algoAC~has two central components, which involve (a) determining whether a state is relevant or not, and (b) selecting actions which will enable the agent to reach and explore relevant states. To tackle (a) we leverage an ensemble of relevance discriminators, and address (b) using a \rev{visual model predictive control (MPC)} approach, both of which we detail next.

\begin{figure}%
    \centering
    \vspace{0.1cm}
    \includegraphics[width=0.99\linewidth]{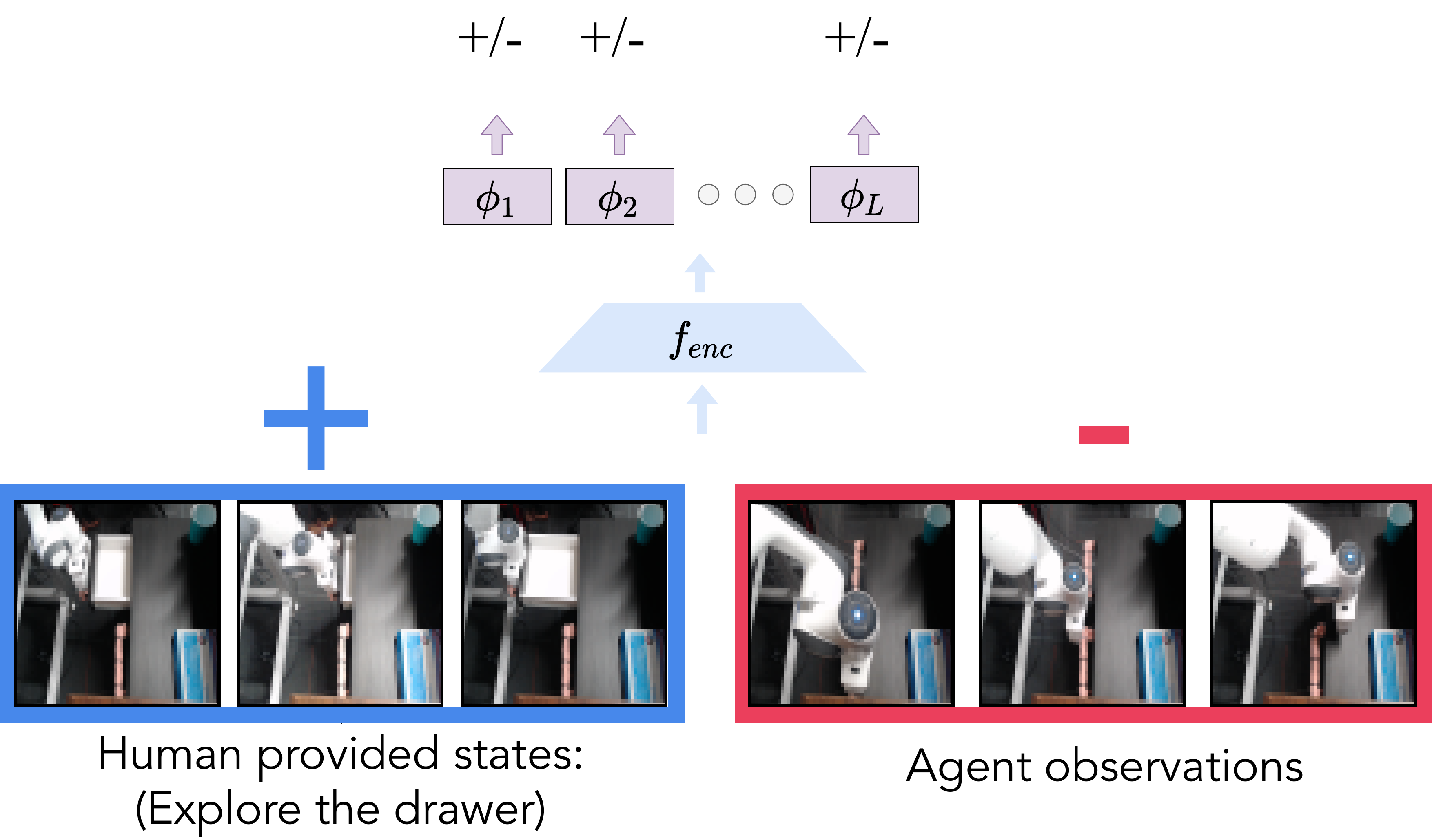}
    \vspace{-0.6cm}
    \caption{\small \textbf{\algoAC~Ensemble of Relevance Discriminators.} \algoAC~learns an ensemble of $L$ discriminators to differentiate between agent states and human provided relevant states. }
    \vspace{-0.6cm}
    \label{fig:disc}
\end{figure}

\begin{figure*}[t]
\vspace{0.1cm}
\centerline{\includegraphics[width=0.99\linewidth]{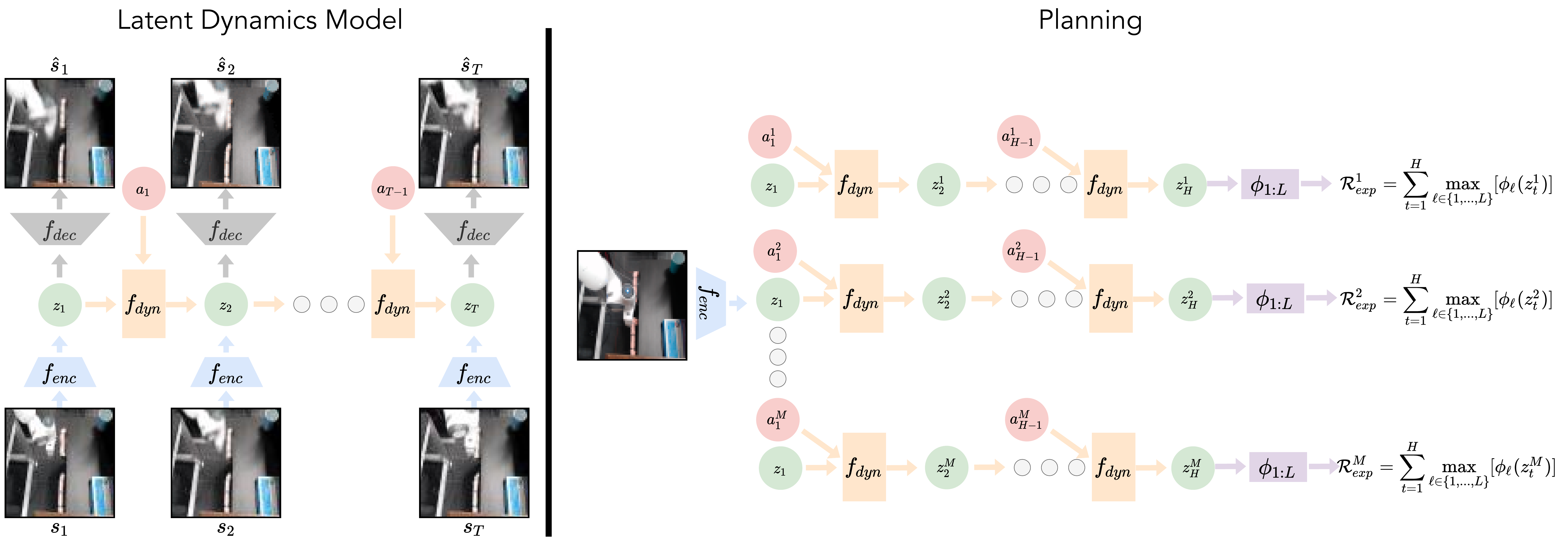}}
\caption{\small{\textbf{Model-Based Planning with \algoAC.} \algoAC~learns a latent dynamics model from pixels (\textbf{left}), then explores with sampling-based planning in the latent space of the learned model to maximize \reward~(\textbf{right}).}}
\vspace{-0.2cm}
\label{inference}
\end{figure*}
\noindent\textbf{Relevance Discriminator.}
To determine which states are relevant, \algoAC~learns an ensemble of $L$ discriminators online which differentiate between the agent's growing dataset \data~of collected experience and the relevant states provided by the human.  Given states $s \sim$ \data~as negatives
and human-provided relevant states $[\bar{s}_1, ..., \bar{s}_K] \sim \mathcal{S}^*$ as positives,
\algoAC~encodes each \rev{into a latent space} using a neural network state encoder $f_{enc}$, then trains each fully connected network $\phi_l$ as a binary classifier for each element of the ensemble, shown in Figure~\ref{fig:disc}. That is, for human examples $s^* \sim [\bar{s}_1, ..., \bar{s}_K]$
and agent experience $s \sim $\data, each discriminator $\phi_l$ is trained
to:
\begin{equation}
     \max_{\phi_l} 
    \mathbb{E}_{s^*}
    [\log(\phi_l (f_{enc}(s^*)))] + \mathbb{E}_{s}
    [\log (1 - \phi_l (f_{enc}(s)))].
    \label{eq:discloss}
\end{equation}
\rev{where $f_{enc}$ is learned with a VAE objective as described in Equation~\ref{eq:vaeloss}.}
Throughout learning, each discriminator in the ensemble is randomly initialized and samples different balanced mini-batches of human states and agent observations.

As identified in prior work \cite{ fu2018variational, singh2019end, torabi2018generative}, a core challenge in training discriminators on a handful of human-provided relevant states and agent observations is when the discriminators may overfit and provide a sparse or difficult to optimize reward signal. This issue is exacerbated in our setting by the relatively small (10-100) number of human examples.
To address this, 
\rev{we use mix-up regulatization \cite{Zhang2018mixupBE} to encourage smooth predictions and to mitigate overfitting we use spectral normalization and data augmentation (random cropping).}

Now that we have described how the discriminators are trained, how do they translate to our exploratory reward \reward? We would like to select action sequences which explore around states for which either the ensemble $[\phi_1, .., \phi_L]$ estimates are relevant, or states for which the ensemble has high uncertainty. 
Therefore, rather than exploring under the mean ensemble score, we use an optimistic estimate given by the maximum discriminator score over the ensemble models: $\text{\reward}(s, \mathcal{C}) = \max [\phi_1({f_{enc}(s)}), ..., \phi_L({f_{enc}(s)})]$, which captures both the predicted relevance and the uncertainty.

\noindent\textbf{Learned Dynamics and Planning.}
To plan actions to maximize the exploratory reward, we take a model-based planning approach, due to its sample efficiency and ability to handle the non-stationary reward \reward. The learned latent dynamics model $p_\theta$ consists of three components, (1) an encoder $f_{enc}(z_t | s_t; \theta_{enc})$ that encodes the state $s_t$ into a latent distribution from which $z_t$ is sampled, (2) a decoder $f_{dec} (s_t | z_t; \theta_{dec}) $ that reconstructs the observation, providing a reconstruction $\hat{s_t}$, and (3) a deterministic forward dynamics model in the latent space $f_{dyn} (\hat{z}_{t+1} | z_t, a_t ; \theta_{dyn})$ which learns to predict the future latent state $z_{t+1}$ from $z_t$ and action $a_t$ (See Figure~\ref{inference}). In our experiments we work in the setting where states are images, so $f_{enc}(z_t | s_t)$ and $f_{dec} (s_t | z_t) $ are convolutional neural networks, and $f_{dyn} (z_{t+1} | z_t, a_t)$ is a recurrent neural network. The encoder and decoder weights $\{\theta_{enc}, \theta_{dec}\}$ are optimized under the standard VAE loss, that is maximizing the lower bound on the likelihood of the data:
\begin{equation}
    \max_{\theta_{enc}, \theta_{dec}} \mathbb{E}_{f_{enc}(z \mid s)} [\log f_{dec} (s \mid z)] - \beta D_{KL}[f_{enc} (z \mid s) || p(z)]. 
    \label{eq:vaeloss}
\end{equation}
The forward dynamics weights $\theta_{dyn}$ are optimized to minimize the mean squared error loss with the next latent state
\begin{equation}
    \min_{\theta_{dyn}}\mathbb{E}_{ f_{enc}(z_{t:t+H} \mid s_{t:t+H})}||z_{t+1:t+H} - f_{dyn}(z_t, a_{t:t+H-1})||^2_2
    \label{eq:dynloss}
\end{equation}
where $H$ indicates the prediction horizon. Exact architecture and training details for all modules are in the supplement.

\algoAC~then uses sampling-based planning, specifically the cross-entropy method (CEM)~\cite{rubinstein2013cross}, in conjunction with this latent dynamics model to plan sequences of actions to maximize the exploratory reward \reward. Concretely, it first encodes its current observation into a latent space $z_t$ using the learned encoder $f_{enc}$. On each iteration of CEM it then samples $M$ action sequences of length $H$, which it feeds through the latent dynamics model $f_{dyn}$, resulting in predicted future states $\hat{z}_{t+1:t+H+1}$, which are ranked according to \reward~(See Figure~\ref{inference}). \rev{Specifically, the reward for the $i^{th}$ sampled trajectory is $\mathcal{R}_{exp}^i = \sum_{{t'}=t+1}^{t+H+1} \max_{\ell \in \{1, ..., L\}} [\phi_\ell(z^i_{t'})]$, that is the maximum discriminator score over the ensemble, summed over the entire predicted trajectory.}

%% file: sections/algoblock.tex
\begin{algorithm}[H]
\caption{$\textsc{\algoAC}([\bar{s}_1, ..., \bar{s}_K]$)}
\begin{algorithmic}[1]
\footnotesize
\STATE \rev{Randomly initialize $\theta, \phi_1, ..., \phi_L$}
\STATE \rev{Initialize \data $\leftarrow \emptyset$}
\STATE \textcolor{blue}{/* \algoAC~reward is max score over discriminators */}
\STATE Let \reward$(s, \mathcal{C}) = \max [\phi_1(f_{enc}({s})), ..., \phi_L(f_{enc}({s}))]$
\FOR{$ep=1, 2, ..., E$}
    \STATE \textcolor{blue}{/* Plan and execute actions under \algoAC~reward */}
    \WHILE{\rev{episode} not done}
    \STATE $z_t \sim f_{enc}(s_t)$
    \STATE $a_{t:t+H} = \textsc{LatentMPC}(z_t, f_{dyn},$ \reward)
    \STATE $s_{t+1:t+H+1} \sim  p( \cdot \mid s_t, a_{t:t+H})$
    \STATE $\mathcal{D} \leftarrow \mathcal{D} ~\bigcup~ \{[(s_t, a_t),..., (s_{t+H+1})] \}$
    \ENDWHILE
    \FOR{num updates $U$}
        \STATE \textcolor{blue}{/* Train dynamics model $p_\theta$ */}
        \STATE $s_{t:t+H}, a_{t:t+H} \sim \mathcal{D}$
        \STATE Update $f_{enc}, f_{dec}$ according to Eq.~\ref{eq:vaeloss}
        \STATE Update $f_{dyn}$ according to Eq.~\ref{eq:dynloss}
        \STATE \textcolor{blue}{/* Train relevance discriminators $\phi$ */}
        \FOR{each $\phi_l$}
            \STATE $s \sim \mathcal{D}, s^* \sim [\bar{s}_1, ..., \bar{s}_K]$
            \STATE Update $\phi_l$ according to Eq.~\ref{eq:discloss}
        \ENDFOR
    \ENDFOR
\ENDFOR
\RETURN dataset $\mathcal{D}$
\end{algorithmic}
\end{algorithm}
\vspace{-0.3cm}

%% file: sections/experiments.tex
\section{Experiments}
\label{exps}

\begin{figure}[b]%
    \centering
    \vspace{-0.3cm}
    \includegraphics[width=\linewidth]{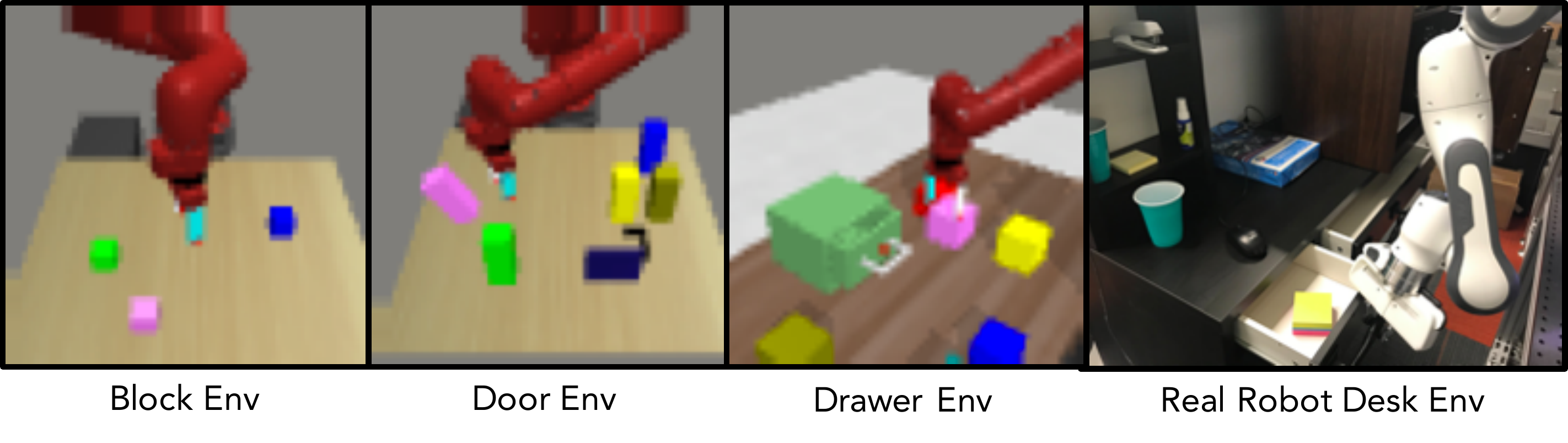}
    \vspace{-0.6cm}
    \caption{\small \textbf{Experimental Domains.} We consider 3 simulated domains, interacting with blocks, a door, and a drawer. We also test on a real Franka robot interacting with a desk.}
    \label{taskex}
\end{figure}

\begin{figure*}[t]
\centerline{\includegraphics[width=0.9\linewidth]{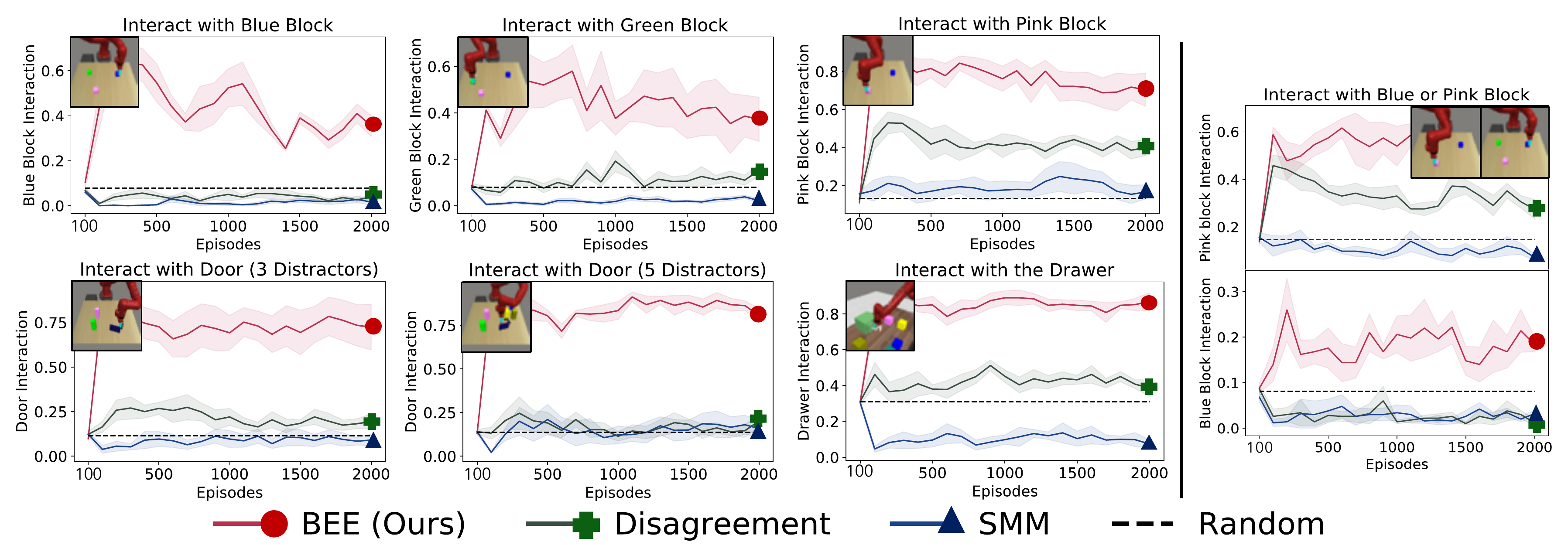}}
\caption{\small{\textbf{Interaction with Target Objects.} Interaction with each target over learning. Specifically, we plot the \rev{relative frequency} of the last 100 episodes where the agent moves the target object more than some fixed threshold (y-axis). For each relevant object, a sample human-provided goal image is shown in the top left. We observe that given a single target object, \emph{ \algoAC~interacts with the target object more than twice as often as the comparisons in all domains} \textbf{(left)}. We also observe that given two targets, \algoAC~is able to effectively explore them both \textbf{(right)}.
} }
\vspace{-0.2cm}
\label{interaction}
\end{figure*}

In our experiments we aim to assess how effectively \algoAC~can explore relevant regions of the state space compared to state-of-the-art approaches in task-agnostic and weakly-supervised exploration. Concretely, we ask the following experimental questions:
\textbf{1)} Does \algoAC~yield improved interaction with relevant objects, while being robust to irrelevant distractor objects?
\textbf{2)} Does using data collected from \algoAC~lead to better downstream task performance than using data collected via state-of-the-art task-agnostic and weakly-supervised exploration techniques?
\textbf{3)} How does \algoAC~perform on hard exploration tasks on a real robotic system from images?
Next, we describe our experimental domains and comparisons in Section~\ref{exp_dom_comp}, then explore the above three questions in Sections \ref{exp1}, \ref{exp2}, and \ref{exp3}. For  video results and supplemental details, see \url{https://sites.google.com/view/batch-exploration}.

\subsection{Experimental Domains and Comparisons.}
\label{exp_dom_comp}

\noindent\textbf{Experimental Domains.} Our experiments focus on tabletop robot manipulation from raw image observations. Specifically, we first consider a suite of hard-exploration manipulation tasks with a simulated Sawyer robot. The simulator is built off of the MetaWorld environment \cite{yu2020meta} and MuJoCo physics engine~\cite{mujoco}, and contains a scene with 3 small blocks, where the exploration objective is to interact with a particular block, a scene with a varying number of distractor towers and a door, where the exploration objective is to interact with the door, a scene with a drawer and a number of distractor blocks, and the exploration objective is to interact with the drawer.
Additionally, we consider a real Franka Emika Panda robot operating over a desk, with the exploration objective of interacting with a small corner drawer, as shown in Figure~\ref{taskex}. 

\noindent\textbf{Comparisons.} We compare \textbf{\algoAC}~to two state-of-the-art exploration methods. First we compare to \textbf{Disagreement}, which uses model disagreement as the exploration objective for planning \cite{pathakICMl17curiosity, sekar2020planning}. This method uses an ensemble of five latent dynamics models, and uses the variance in their predictions as the reward for planning, similar to \cite{sekar2020planning}.  
Second, we consider state marginal matching \cite{lee2019efficient} (\textbf{SMM}), which also uses the human-provided weak supervision.
Specifically, it fits a density model to the human-provided relevant states $p^*(s)$, and plans under the standard SMM reward $\mathcal{R}(s) = \log p^*(s) - \log p(s)$ where $p(s)$ is the distribution over the policy's visited states. Lastly, we compare to \textbf{Random}, a completely random exploration policy.
\rev{In all experiments we used the default hyperparameters for regularization like mix-up, cropping, and spectral normalization for
\algoAC~and comparisons, and found them to work effectively in both simulated and real experiments without further tuning, suggesting that the method is robust to these hyperparameters.}
For more implementation details, please refer to Appendix~\ref{app:tr}. 

\subsection{Does \algoAC~Interact More With Relevant Objects?}
\label{exp1}

We begin by measuring how much \algoAC~and the comparisons
interact with relevant objects specified by the human,
shown in Figure~\ref{interaction}. We include guided exploration (100 human provided images) toward each of the 3 small blocks, as well as to the door under varying numbers of distractors, and to the drawer. We report the percentage of episodes in which the agent moved the target object more than a threshold every 100 episodes,
with the first 100 episodes corresponding to random interaction. We observe that over all domains and targets \algoAC~interacts with the relevant object more than twice as often as the prior methods (Figure~\ref{interaction} (left)). Furthermore, we observe that \algoAC~can be used to guide exploration towards not just a single object, but multiple target objects, and find that \algoAC~interacts with both more than the other methods (Figure~\ref{interaction} (right)).
We also observe that the model disagreement method explores more effectively than random; but when there are many distractors, it interacts more with the distractors than with the relevant object. SMM exhibits poor performance likely due to the learned density models struggling to scale to high dimensional image observations.

 \begin{figure*}
\vspace{0.1cm}
\centerline{\includegraphics[width=0.68\linewidth]{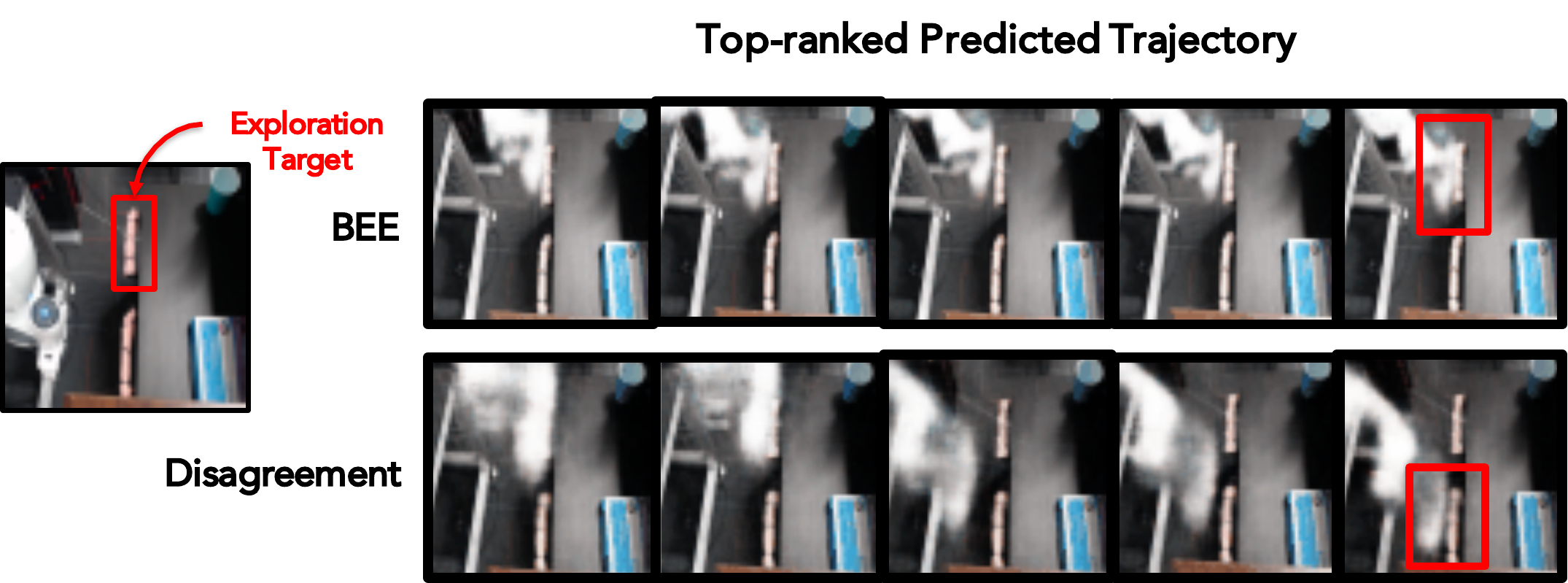}}
\caption{\small{Examples of predicted trajectories that are ranked high during online data collection on the robot for BEE vs. Disagreement. BEE rankings effectively discriminate target and non-target interaction, with the high-ranked trajectory under BEE going to the target object (corner drawer) even in the earlier steps of the episode. High-ranked states under Disagreement do not involve the target object.} }
\vspace{-0.2cm}
\label{rankings_robot}
\end{figure*}

\subsection{Does \algoAC~Data Enable Better Downstream Performance?}
\label{exp2}
For downstream batch RL using the collected data, we consider the model-based self-supervised RL setting. Specifically, we consider the visual foresight algorithm \cite{finn2017deep, ebert2018visual}, which learns a model of the dynamics from an unlabeled batch of interaction data, then uses this model with planning to reach goals. We train a visual dynamics model using Stochastic Variational Video Prediction (SV2P) \cite{babaeizadeh2017stochastic}. The model is trained on the full dataset \data~collected in the batch exploration phase to predict future states, i.e. $(s_{t+1:t+H} | s_t, a_{t:t+H-1})$. The architecture and losses used are identical to the original SV2P paper \cite{babaeizadeh2017stochastic}, and can be found in the appendix. For downstream task planning, we use the SV2P model with sampling based planning (CEM), to plan sequences of actions to reach a goal image, under the planning cost of $\ell_2$ pixel distance.
\begin{table}
  \small{
  \centering
  \def\arraystretch{0.95}
  \begin{tabular}{cccccc}
  \toprule
  \!\!\ \!\! & \!\! Open \!\! &  \!\! Push \!\! &  \!\! Push \!\!  &  \!\! Push \!\!  &  \!\! Push \!\!   \\
  
  \!\!\ \!\! & \!\! Drawer \!\! &  \!\! Door (3) \!\! &  \!\! Door (5) \!\!  &  \!\! Green \!\!  &  \!\! Blue \!\!   \\
  \midrule  %
  \!\!\algoAC~(Ours)\!\! & \!\!\textbf{ 0.42} \!\! & \!\! \textbf{0.63} \!\! & \!\! 0.65 \!\!  & \!\! \textbf{0.47} \!\!  & \!\! \textbf{0.50} \!\!  \\ 
  \hline
    \!\!Disagreement\!\!  & \!\! 0.36 \!\! & \!\! 0.59 \!\!  & \!\! 0.69 \!\!   & \!\! 0.45 \!\!  & \!\!  0.44 \!\!  \\ 
  \hline
  \!\!SMM\!\!  & \!\! 0.29 \!\! & \!\! 0.58 \!\!  & \!\! \textbf{0.70} \!\!  & \!\! 0.43 \!\!  & \!\! 0.46 \!\!  \\ 
  \hline
  \!\!Random\!\!  & \!\! 0.31 \!\! & \!\! 0.60 \!\!  & \!\! 0.65 \!\!  & \!\! 0.45 \!\!  & \!\! 0.45 \!\!  \\ 
  \bottomrule
 \end{tabular}
 }
 \vspace{0.1cm}
  \caption{\small{\textbf{Downstream success rates using planning with collected data. } We compare the downstream task performance of using the data generated by \algoAC~for batch RL using the visual foresight method. We observe that across 4 out of 5 tasks \algoAC~is the top performing method. All results are averaged over 1000 trials. 
  } }
  \vspace{-0.6cm}
  \label{tab:success}
\end{table}

We consider five downstream planning tasks: (a) opening the drawer,
(b) pushing the door with  3 distractors,  (c) pushing the door with 5 distractors, (d) pushing the green block left, and (e) pushing the blue block right. We observe in Table \ref{tab:success} that across 4 out of 5 tasks, using the data from \algoAC~improves performance over prior methods and random exploration.

\subsection{Is \algoAC~Effective on a Real Robot?}
\label{exp3}

\begin{figure}
\centerline{\includegraphics[width=0.9\linewidth]{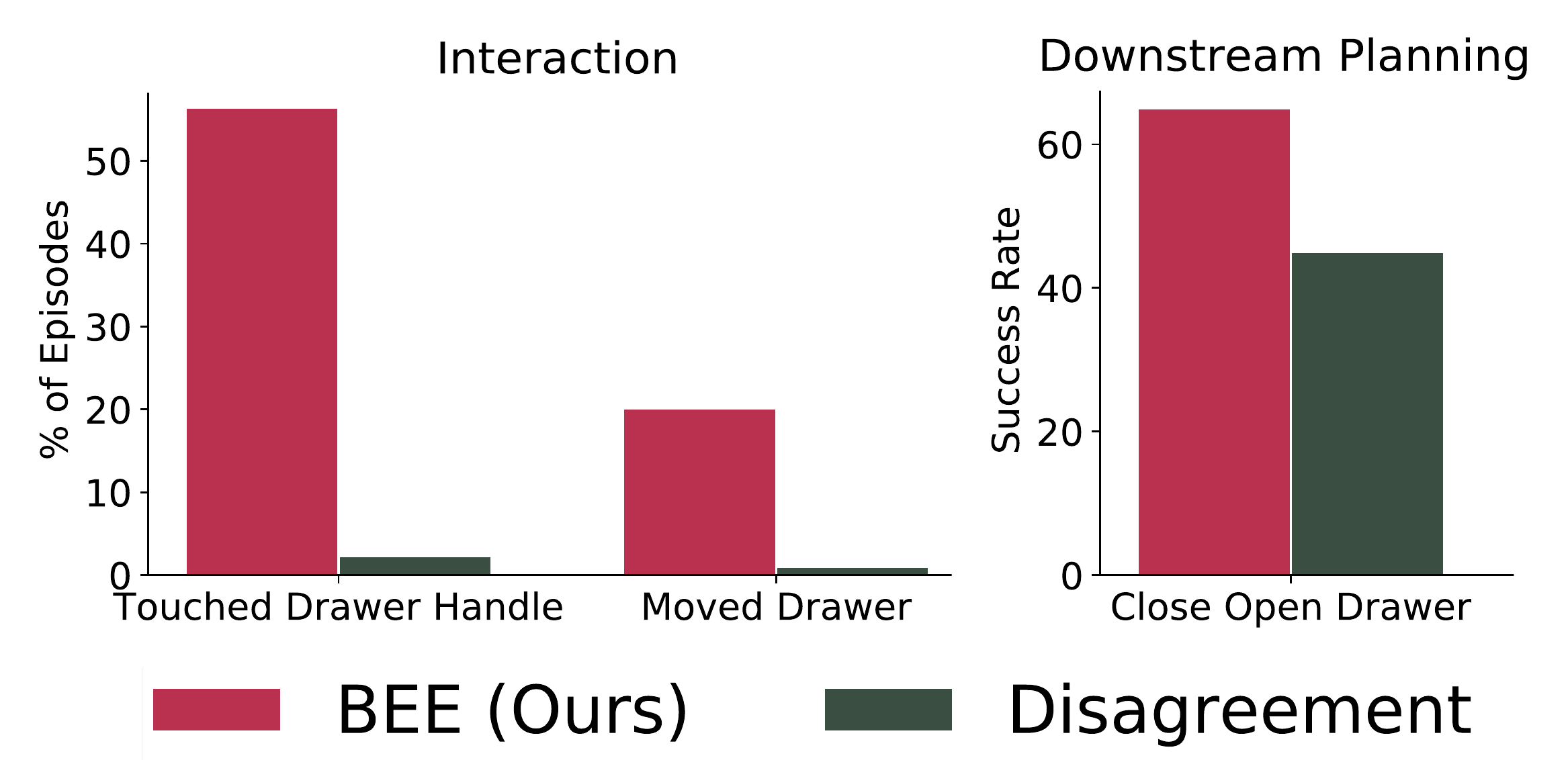}}
\caption{\small{\textbf{Performance on a Real Robot.} On the desk interaction task with a real Franka robot, we report the percentage of episodes in which the agent interacts with the target drawer (left, middle), as well as the success rate over 20 trials in the downstream task of closing the drawer (right). We observe that \algoAC~interacts an order of magnitude more with the drawer than Disagreement, and yields a 20\% improvement in downstream planning.
  }}
\vspace{-0.5cm}
\label{robot_res}
\end{figure}

Next, we test if \algoAC~can tackle challenging exploration problems on a real robot from pixel inputs. To do so, we consider the domain of a Franka Emika Panda robot positioned over a desk, shown in Figure~\ref{taskex}. 
We provide weak supervision (50 human provided images)
encouraging interaction with the smaller of the two drawers, and measure the extent
to which \algoAC~interacts with the target object compared to Disagreement. Note that none of the algorithms have access to this measurement, and it is computed after data collection by visual inspection. As shown in Figure~\ref{rankings_robot}, \algoAC~highly ranks trajectories that interact with or around the drawer, while Disagreement does not, resulting in better exploration around the drawer. As we see in Figure~\ref{robot_res}, not only does \algoAC~interact with the drawer an order of magnitude more than Disagreement, but a dynamics model learned on the collected data is 20\%
more successful for the downstream task of closing the drawer. For downstream planning, we again use planning with a video prediction model trained on each methods data, %
leveraging a reward model trained on a few hundred labeled examples of a closed drawer. 
\revf{Despite being less accurate at modeling the drawer, the Disagreement model was able to get reward by moving its arm in the right direction, and thus occasionally closed the drawer.}

\subsection{Ablating \algoAC~Design Choices}
\rev{
Lastly, we ablate various design choices in \algoAC~and their effect on exploration performance. In Figure~\ref{ab1} of the appendix we examine how the choice of using the maximum discriminator score compares to using (a) a sum of mean and variance,
and (b) using just a single discriminator. We find that max performs better than the combined mean and variance, and 
that while on average a single discriminator also performs well, it is far more unstable and prone to sustained drops in performance. Additionally, in Figure~\ref{ab2} in the appendix we study the effect of the hidden dimension size, and find that BEE is robust to this hyperparameter.
}
\label{exp3}

%% file: sections/future_work.tex
\section{Limitations and Future Work}

We have presented batch exploration with examples (\algoAC) as a technique for scalable collection of diverse robotic datasets which guides exploration with weak human supervision. While we observe increased interaction with target objects and improved downstream performance in simulated and real robot domains, limitations remain. First, a core aspect of \algoAC~is leveraging weak human supervision in the form of a handful of "relevant states" to explore. However, how exactly these states are selected can potentially have a large impact on the quality of exploration. Specifically, if the distribution of states is too narrow, \algoAC~may simply visit only those exact states. On the other hand, if the distribution of states is too broad, \algoAC~may only explore some subset of it. While we found \algoAC~to be generally insensitive to this choice, studying more closely how best to generate the human supervision for \algoAC~is an interesting direction for future work.
\rev{Second, as is the case with any visual RL algorithm deployed on a real robot, experimental choices are important to successful robotic exploration and data collection, such as choice of camera viewpoint, action space,
or the environment's dependence on 
human resets (e.g. in the case of knocking objects off the table). Addressing these challenges is important for enabling scalable robotic data collection for a breadth of tasks. }
Despite these challenges, \algoAC~provides a step towards enabling the collection of larger and more diverse robotic datasets, which may be key to learning general purpose robotic agents.

%% file: sections/appendix.tex
\section{Training Details}
\label{app:tr}

\textbf{Acquiring human supervision:} For each comparison in each simulated domain, we supply 100 examples of relevant images. For the block domain, these examples involve the gripper hovering over the target block at a random z position in a region of +/- 2 cm in the x and y directions from the initial block position. For the door domain, the example images involve the gripper next to the door handle with the door set to random angles either between -45 and -5 degrees or between 5 and 45 degrees. For the drawer domain, the example images involve the gripper near the drawer handle, with the drawer open to random amounts between 0 and 14 cm. For the robot domain, the example images involve the small corner drawer of the desk opened and the robot arm moved to the handle. For each comparisons in the robot domain, we only supply 50 examples of relevant images.

\textbf{Planning during online data collection (MPC):} For all simulation domains, we report results averaged over 5 different random seeds for each method. For each run, we collect a dataset of 2,000 episodes, each of 50 time steps. During online planning, all methods use a single iteration of the cross entropy method to plan a sequence of actions. 
For each 50-step episode, we replan every 10 steps, i.e. we plan five 10-step trajectories. In simulation, each episode is reset to a fixed initial state. For the real robot domain, we run one random seed for 1000 episodes, each of 100 time steps, also replanning every 10 steps.
At each stage of planning, we sample 1000 10-step action sequences and sort according to the method used. The agent uniformly randomly chooses one of the top 5 ranked trajectories to execute. With probability 0.1, the agent takes a random action in place of a chosen one from the selected trajectory. 

\textbf{Model training:} All models for BEE, Disagreement, and SMM are trained with a learning rate of 1e-3. The main VAE ($f_{enc}, f_{dec})$ for all methods uses a beta of 1e-3, and the separate VAEs for SMM use beta 0.5, which was the default value used in the codebase of the original paper. After each new episode is collected, it is added as a sample of size [50, 3, 64, 64] into the replay buffer. The encoder/decoder $f_{enc}$ and $f_{dec}$ as well as the dynamics model (all 5 in the case of Disagreement) are updated 20 times after each new episode.  For SMM, both VAEs are also updated 20 times after each epoch. All models are trained using separate Adam optimizers and using random batches of size 32 length H samples starting from any time step of the most recent 500 episodes, where H is the current training horizon for the dynamics model(s). 

For training the dynamics model(s), for the first 50 episodes, we use a training horizon of 2; for the next 100 episodes, we use horizon 4; for the 150 episodes after that, we use horizon 8; and for all remaining episodes we use horizon 10. For all comparisons, the encoder/decoder $f_{enc}$ and $f_{dec}$ are updated for each of the 20 times with 1 batch from observations in the replay buffer that were collected online as well as 1 batch from the provided example images. Cropping regularization is applied to these input batches by expanding the boundaries by 4 pixels each and then choosing a random $64 \times 64$ crop of this larger image. For all simulated experiments, balanced batches of both human-provided example images and observations from the replay buffer are used to train the main VAE. Hence, all methods in simulation leverage the same human-provided weak supervision. In the robot domain (for all methods), the VAE is not trained with balanced batches of the human provided images, as there are only a small number (50) such states.

For BEE, the relevance discriminators are each updated once at the end of the episode. To prevent overfitting, the discriminators are trained with mixup and input image cropping. For mixup regularization, hyperparameter $\alpha = 1$ is used to control the extent of mixup.

\section{Experimental Details}
\label{app:ex}

For the block, door, and drawer domains, we use a Mujoco simulation built off the Meta-World environments \cite{yu2020meta}. For the robot domain, we consider a real Franka robot operating over a desk, which has two drawers as well as a cabinet and multiple objects on top. The state space is the space of RGB image observations with size [64, 64, 3]. For the simulation env, we use a continuous action space over the linear and angular velocity of the robot’s gripper and a discrete action space over the gripper open/close action, for a total of five dimensions. 

\textbf{Interaction with Target:}
For the block and door evaluation of the online data collection, interaction is defined as moving the target block or door at least a distance of 5 cm any time during the episode. For the drawer domain, interaction is defined as pulling the drawer open by at least 3 cm any time during the episode. The drawer begins slightly open (by 5 cm distance). Lastly, for the real robot domain, we define two criteria for interaction: (1) touching the handle of the desk's corner drawer and (2) actually moving the drawer open or closed. We do not reset the drawer position between episodes, so if an episode ends with the drawer open, the next episode will start with it open. 

\textbf{Downstream Planning:}
For all control experiments, evaluation is done by using model predictive control with SV2P models trained on the full datasets collected from each of five seeds in the batch exploration phase (a total of 10k episodes) along with 5k random episodes for 100k iterations. For evaluating control on the real robot, for each method we train the SV2P model on the 1000 episodes collected in the batch exploration phase and no random data. We plan 10 actions and execute them in the environment five times for a 100 step trial. Each stage of planning uses the cross entropy method with two iterations, sampling 200 10-step action sequences, sorting them by the mean pixel distance between the goal and the predicted last state of each trajectory, refitting to the top 40, and selecting the lowest cost trajectory. 

\begin{figure}[h!]
\centerline{\includegraphics[width=\linewidth]{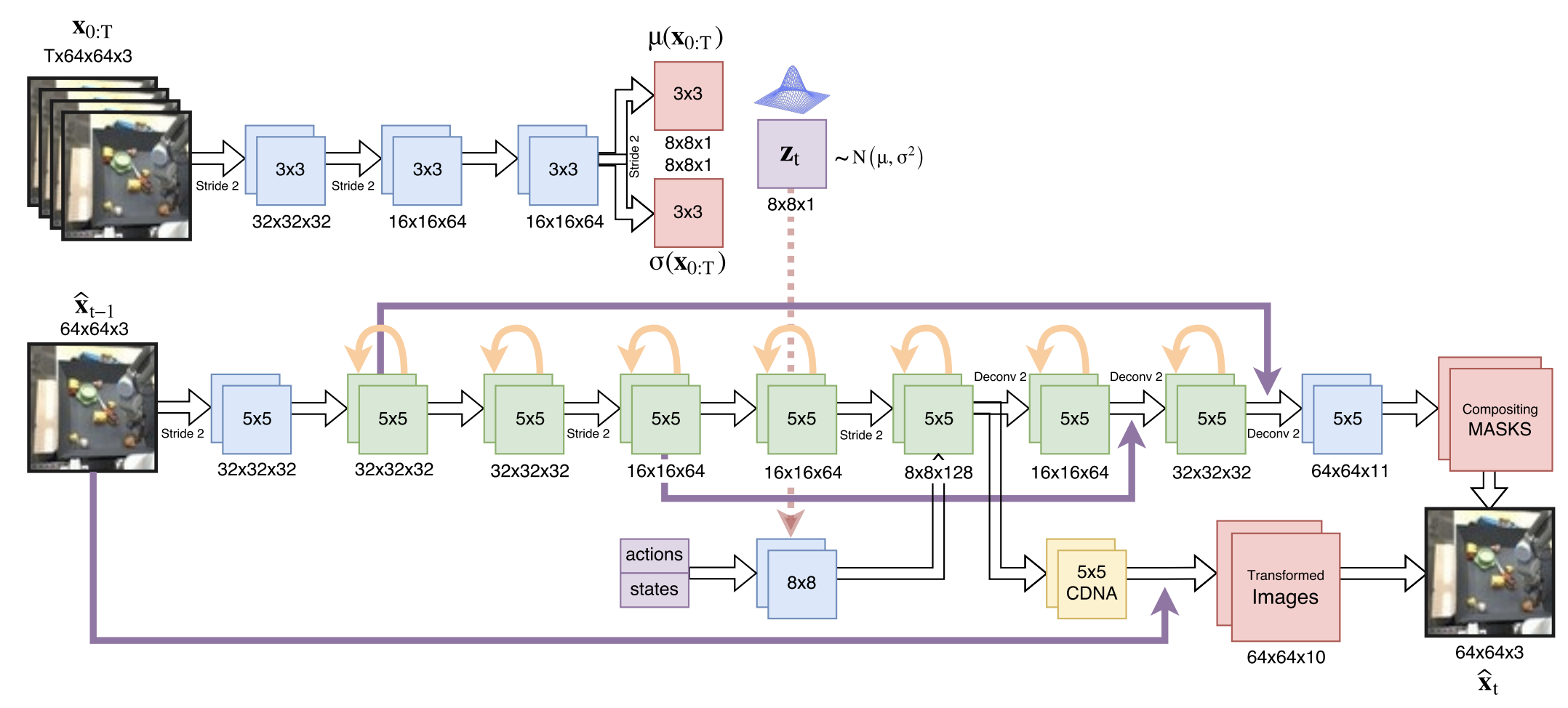}}
\caption{\small{\textbf{SV2P architecture.} SV2P estimates the posterior latent distribution $p(z \mid x_{0:T})$ by learning an inference network (top) $q_{\phi}(z \mid x_{0:T}) = \mathcal{N}(\mu(x_{0:T}), \sigma(x_{0:T}))$. Latent values are sampled from $q_{\phi}(z \mid x_{0:T})$, and the generative network (bottom) takes in the previous frames, latent values, and actions to predict the next frames. Figure taken from the original paper \cite{babaeizadeh2017stochastic}. 
} }
\label{sv2p}
\end{figure}

\textbf{SV2P Training:} SV2P learns an action-conditioned video prediction model by sampling a latent variable and subsequently generating an image prediction with that sample.  The architecture and losses used here are identical to the original SV2P paper \cite{babaeizadeh2017stochastic}. This architecture is shown in Figure~\ref{sv2p}, which is taken from the original paper. The models are trained to predict the next fifteen frames given an input of five frames. All other hyperparameters used for training are default values used in the codebase of the original paper.

\textbf{Downstream Task Evaluation:}
In the Open Drawer task, the goal image involves the gripper above the drawer handle, which is open to 15 cm distance.  Success is defined by opening the drawer at least 3 cm.
In the Blue Block task, the goal image involves the gripper over the initial blue block position and the blue block moved 10 cm to the right. Success is defined as pushing the block more than 5 cm to the right. 
In the Green Block task, the goal image involves the gripper over the initial green block position and the green block moved 10 cm to the left and 0.04 downwards. Success is defined as pushing the block more than 0.05 to the left. 
In the Door task with five distractors, the goal image involves the gripper above the handle and the door opened to 0.35 radians. Success is defined by opening the door to at least 0.15 radians, measured at the end of each 50-step episode.
For the robot downstream task, we labeled the data collected by BEE and Disagreement and trained a reward classifier on 100 examples (labeled as 1) of the drawer open and 200 examples of the drawer closed (labeled as 0), with the gripper sometimes but not always near the drawer. We then conducted planning using this same classifier as the cost for both methods. Success is defined as pushing the drawer closed.

\section{Architecture Details}
\label{app:a}

In this section, we go over implementation details for our
method as well as our comparisons.

During data collection, for each domain (block, door, and drawer domains in simulation as well as the real robot domain), all comparisons are trained on an Nvidia 2080 RTX, and all input observations are [64, 64, 3]. Each domain  leverages an identical architecture, which is described as follows. 

All comparisons use an encoder $f_{enc}$ with convolutional layers (channels, kernel size, stride): [(32, 4, 2), (32, 3, 1), (64, 4, 2), (64, 3, 1), (128, 4, 2), (128, 3, 1), (256, 4, 2), (256, 3, 1)] followed by fully connected layers of size [512, 2$\times L$] where $L$ is the size of the latent space (mean and variance). We use a latent space size of 256. All layers except the final are followed by ReLU activation.

The decoder $f_{dec}$ takes a sample from the latent space of size $L$ and feeds it through fully connected layers [128, 128, 128], followed by de-convolutional layers (channels, kernel size, stride): [(128, 5, 2), (64, 5, 2), (32, 6, 2), (3, 6, 2)]. All layers are followed by ReLU activation except the final layer, which is followed by a Sigmoid.

The dynamics model $f_{dyn}$ is an LSTM layer [128] followed by a fully connected network with layers [128, 128, 128, L], which are all followed by
ReLU activation except the final layer. For all domains, BEE and SMM learn just one dynamics model while Disagreement learns five of these.

For BEE, we learn an ensemble of three relevance discriminators. These take a sample from the latent space of size $L$ and feed it through fully connected layers [128, 64, 64, 1]. We apply spectral normalization after each layer followed by a ReLU activation (except the final layer, which is followed by a Sigmoid instead). 

For SMM, we learn two separate VAEs: one to represent the density over the policy's visited states while the other fits a density model to the human provided relevant states. These two VAEs have the same architecture: they both use an encoder $g_{enc}$ that takes in a sample from the latent space of size $L$ and feeds it through fully connected layers [150, 150], which are followed by ReLU activations. This is followed by a fully connected layer [$L_2$] for the mean and variance each, where $L_2$ is the size of the latent space. We use $L_2 = 100$. The decoder $g_{dec}$ takes in a sample from the latent space of size $L_2$ and feeds it through fully connected layers [150, 150, $L$], where all layers except the last are followed by a ReLU activation.

\rev{
\section{Additional Results}}

 \begin{figure}
\centerline{\includegraphics[width=0.99\linewidth]{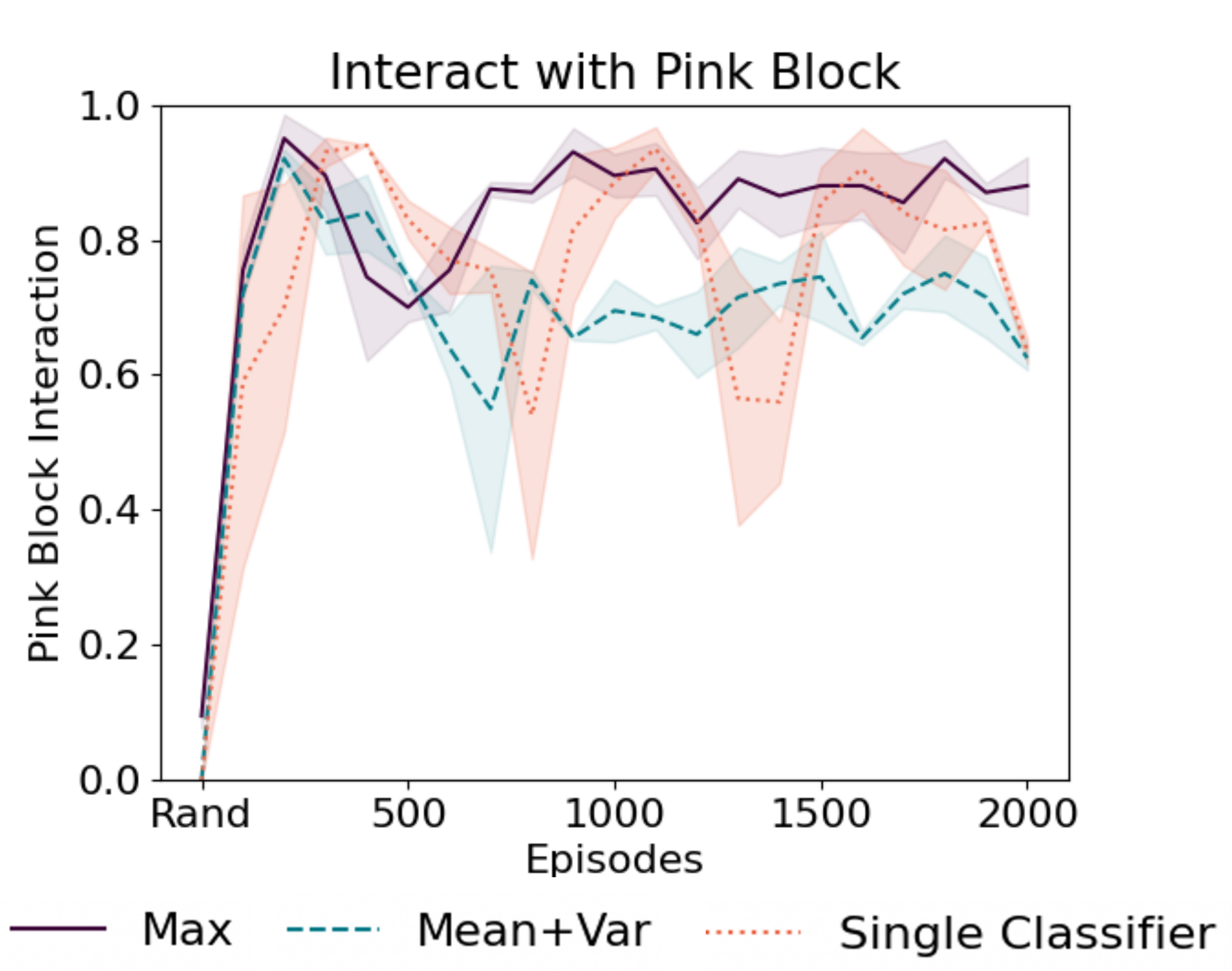}}
\caption{\small{\rev{\textbf{Ablation 1: Different Exploratory Rewards.} We compare different exploratory reward choices in the simulated environment guiding exploration towards the pink block. We find that the max over discriminators which we use performs better than equally weighting mean and variance over the ensemble, and that while using a single classifier also performs well on average, it is naturally more unstable and prone to larger dips in performance. }
  }}
\vspace{-0.5cm}
\label{ab1}
\end{figure}

\begin{figure}
\centerline{\includegraphics[width=0.99\linewidth]{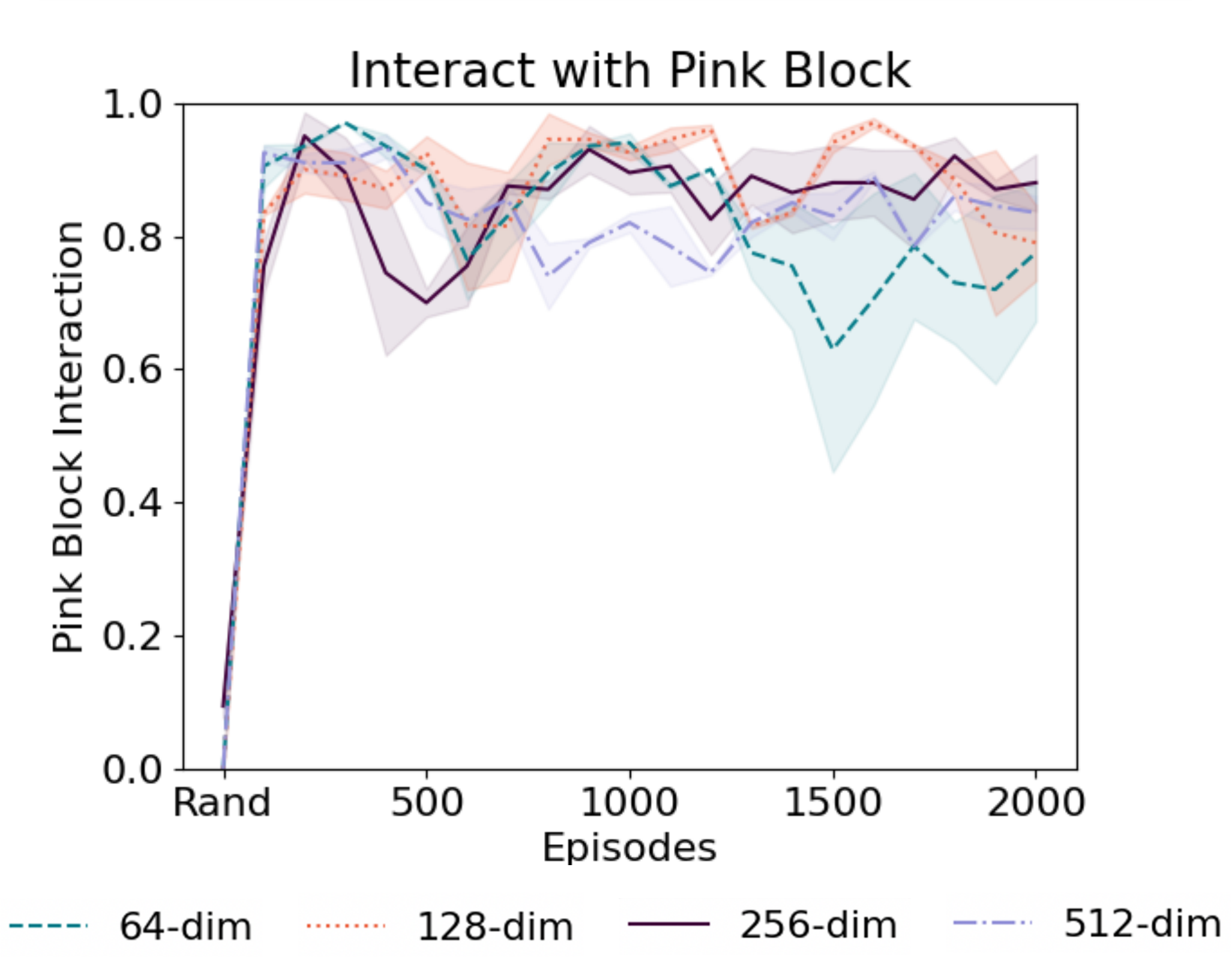}}
\caption{\small{\rev{\textbf{Ablation 2: Various Latent Space Dimensions.} We investigate the sensitivity of BEE to a key hyperparameter, the dimension of the latent space. We find that for the simulated pink block environment, interaction performance with BEE is generally robust across four different latent space dimensions.}
  }}
\vspace{-0.3cm}
\label{ab2}
\end{figure}

\rev{
Here we include plots for our ablation experiments.
First, in Figure~\ref{ab1} we study how \algoAC~which uses the max over the ensemble of discriminators compares to an equally weighted sum of the mean and variance score of the ensemble, as well as using a single discriminator instead of an ensemble. We find that using the max operator in \algoAC~performs comparably or better to combining the mean and variance over the ensemble. Additionally, while just a single classifier also performs comparably on average, it suffers from instability and occasionally has large dips in performance, which are undesirable for unsupervised data collection, as they result in large amounts of useless data. Each method is run over 2 seeds, displaying mean and standard error shading.}

\rev{
Second, in Figure~\ref{ab2} we study \algoAC's robustness to the size of the hidden dimension used for the latent space. This latent space is used for forward dynamics modeling and for learning the relevance discriminators. We observe in Figure~\ref{ab2} that \algoAC~is robust to this choice, and performs effectively for a large range of hidden dimension sizes. Each latent dimension is run over 2 seeds, displaying mean and standard error shading.}